# Augmenting Rating-Scale Measures with Text-Derived Items Using the Information-Determined Scoring (IDS) Framework

**Joe Watson[1,2]*, Ivan O'Connor[3], Chia-Wen Chen[1], Luning Sun[1], Fang Luo[4], David Stillwell[1]**

[1]Psychometrics Centre, Judge Business School, University of Cambridge, Cambridge, UK

[2]Faculty of Law, University of Cambridge, Cambridge, UK

[3]School of Computer Science, University of Birmingham, Birmingham, UK

[4]Faculty of Psychology, Beijing Normal University, Beijing, China

* Corresponding author: jmw239@cam.ac.uk

# Abstract

Psychological assessments commonly rely on rating-scale items, which require respondents to condense complex experiences into predefined categories. Although rich, unstructured text is often captured alongside these scales, it rarely contributes to measuring the target trait because it lacks direct mapping to the latent scale. We introduce the Information-Determined Scoring (IDS) framework, where large language models (LLMs) score free-text responses with simple prompts to generate candidate items that are co-calibrated with a baseline scale and retained based on the psychometric information they provide about the target trait. This marks a conceptual departure from traditional automated text scoring by prioritising information gain over fidelity to expert rubrics or human-annotated data. Using depression as a case study, we developed and tested the method in upper-secondary students (n = 693) and a matched synthetic dataset (n = 3,000). Across held-out test sets, augmenting a 19-item rating-scale measure with LLM-derived items yielded significant improvements in measurement precision and accuracy, and stronger convergent validity with an external suicidality measure throughout the adaptive test. In adaptive simulations, LLM-derived items contributed information equivalent to adding up to 6.3 and 16.0 rating-scale items in real and synthetic data, respectively. This enabled earlier high-precision measurement: after 10 items, 46.3% of respondents reached SE ≤ .3 under the strongest augmented test compared with 35.5% at baseline in real data, and 60.4% versus 34.7% in synthetic data. These findings illustrate how the IDS framework leverages unstructured text to enhance existing psychological measures, with applications in clinical health and beyond.

# Keywords

# Author Note

This research received no specific grant from any funding agency.

The authors declare no conflicts of interest.

Early versions of this work were presented by Joe Watson at the following conferences: *Frontier Research in Educational Measurement (FREMO)*, Oslo, Norway (2025); *International Conference on Cognitive and Psychological Assessment and Enhancement (ICCPAE)*, Beijing, China (2025); and the *9th Conference of the International Association for Computerized Adaptive Testing (IACAT)*, Seoul, Korea (2025). Additionally, a preprint of this manuscript is available on arXiv at https://arxiv.org/abs/2510.08663.

# Introduction

Constructs such as depression are commonly measured in psychological science using established rating scales. These instruments require respondents to condense complex thoughts, feelings, and experiences into predefined response categories. In doing so, this measurement format necessarily compresses richer expressions that respondents may otherwise convey through their primary form of communication – natural language (Kjell et al., 2024). This study introduces the Information-Determined Scoring (IDS) framework, a method for extracting measurement information from free-text responses on the same latent construct as an existing rating-scale test. The framework applies multiple large language model (LLM) scoring methods to each free-text response, generating a set of candidate LLM items. These candidate LLM items are co-calibrated with the existing rating scale and evaluated according to the psychometric information they provide about the target trait. Those whose scoring methods yield the greatest information are retained as LLM items used to augment the original assessment. Using depression as an illustrative case, we show that this integration yields substantial gains in measurement precision and accuracy.

The integration of rating scales with text-derived measurement is relevant in many research settings where structured scale responses are collected alongside open-ended language. The present case study provides a direct application to the measurement of depression, a context in which best-practice clinical assessment already combines standardised questionnaires with open-ended interviews (American Psychiatric Association, 2015; National Institute for Health and Care Excellence, 2022). More generally within clinical psychology, large volumes of language data are increasingly recorded in formats such as therapy transcripts, reflective journals, and voice diaries, creating opportunities to integrate these qualitative responses with established scales. Similar mixed-format measurement designs appear in organisational research, where constructs such as leadership or workplace wellbeing are often assessed using rating-scale batteries alongside open-ended self-evaluations. Beyond psychology, such designs are also common in adjacent domains that rely on latent constructs measured using mixed-format data. For example, structured ratings can be combined with free-text reviews to refine measures of seller proficiency in online marketplaces, or with open-ended responses from large-scale social science surveys – such as the American National Election Studies (ANES) – to improve measurement of constructs like governmental trust (*SI 1*, American National Election Studies, 2025).

Our framework builds on work in automatic essay scoring (AES), a class of computer-based systems for evaluating qualitative responses (Ramesh & Sanampudi, 2022). Early AES systems relied on expert-defined rubrics and rule-based algorithms, reflecting the assumption that complex traits could be adequately represented by finite, hand-engineered feature sets. A canonical example is Page's Project Essay Grade (PEG) system, which quantified writing quality using surface-level "proxes" such as word length and sentence complexity (Ifenthaler, 2022; Maris & Bechger, 2006; Page, 1966). By the 2010s, the field had increasingly adopted machine learning and transformer-based models to predict human judgments. In educational contexts, these models were trained to predict the scores assigned by human raters (Ifenthaler, 2022), while in psychological applications, transformer-derived features were used to classify traits such as politeness (Ludwig et al., 2021). This use of the transformer architecture has now shifted from employing its features for classification to exploiting LLMs directly as scoring mechanisms.

LLMs have advanced both the performance and accessibility of AES systems, but the dominant paradigm remains focused on replicating expert human judgment. This typically requires one of two forms of human input: a detailed, predefined scoring rubric, or a large set of high-quality, human-labelled training data. In educational settings, for instance, LLM-based AES has been used to apply human-written rubrics to German essays and physics responses (Kortemeyer & Nöhl, 2025; Seßler et al., 2025), albeit with such applications requiring human oversight (Kortemeyer & Nöhl, 2025). Similarly, in psychological research, LLM approaches can accurately predict human ratings of mental health conditions (Kermani et al., 2025; Qiu et al., 2025), given the availability of sufficient human-annotated training data. The prevailing approach is therefore limited not just by the practical scaling challenges of sourcing expert input (Kumar et al., 2024; Maris & Bechger, 2006), but by its conceptual tether to replicating a predefined "correct" way of scoring.

While researchers have begun to integrate LLM-scored text with psychometric scales, many existing approaches stop short of placing these scores within a unified measurement model. For example, some analyse LLM-scored text in isolation, either fitting unidimensional models to rubric-derived scores (Kortemeyer & Nöhl, 2025) or organising them into novel factor structures without validating them against established scales (Simons et al., 2024). Others compare AI-scored text directly with human-completed surveys (albeit identifying only weak correlations with established traits: Peters & Matz, 2024), or use LLMs to populate responses to an established depression scale from free-text inputs (Ravenda et al., 2025). Our approach builds on efforts such as the Duolingo English Test (Cardwell et al., 2023), which integrates LLM-scored language responses with closed-ended items within a unified IRT model so that both contribute to measurement on the same latent scale. The IDS framework extends this idea. Rather than relying on a single expert-created scoring rubric or human-annotated data, IDS evaluates multiple candidate scoring methods for the same text responses and uses their co-calibration with an existing rating scale to determine which method should be retained. Co-calibration is therefore used not only to place text-derived scores on the same latent scale, but also to determine how the text should be scored in the first place.

The IDS framework proceeds through four stages. As illustrated in Figure 1, the framework takes as inputs responses to a calibrated rating-scale test together with free-text responses from the same individuals (Stage 1). A pool of candidate LLM-derived items is then generated by applying simple scoring prompts to the text (Stage 2). Each candidate item is subsequently evaluated by co-calibrating it with the baseline scale and quantifying the psychometric information it provides about the target trait (Stage 3). Finally, the items that contribute the greatest information are selected to construct one or more augmented tests (Stage 4).

We evaluated the proposed framework using the following hypotheses:

H1: An augmented test will yield trait estimates that are more precise (have a lower standard error) than a baseline rating-scale-only test.

H2: An augmented test will yield trait estimates that are more accurate (closer to true theta) than a baseline rating-scale only test.

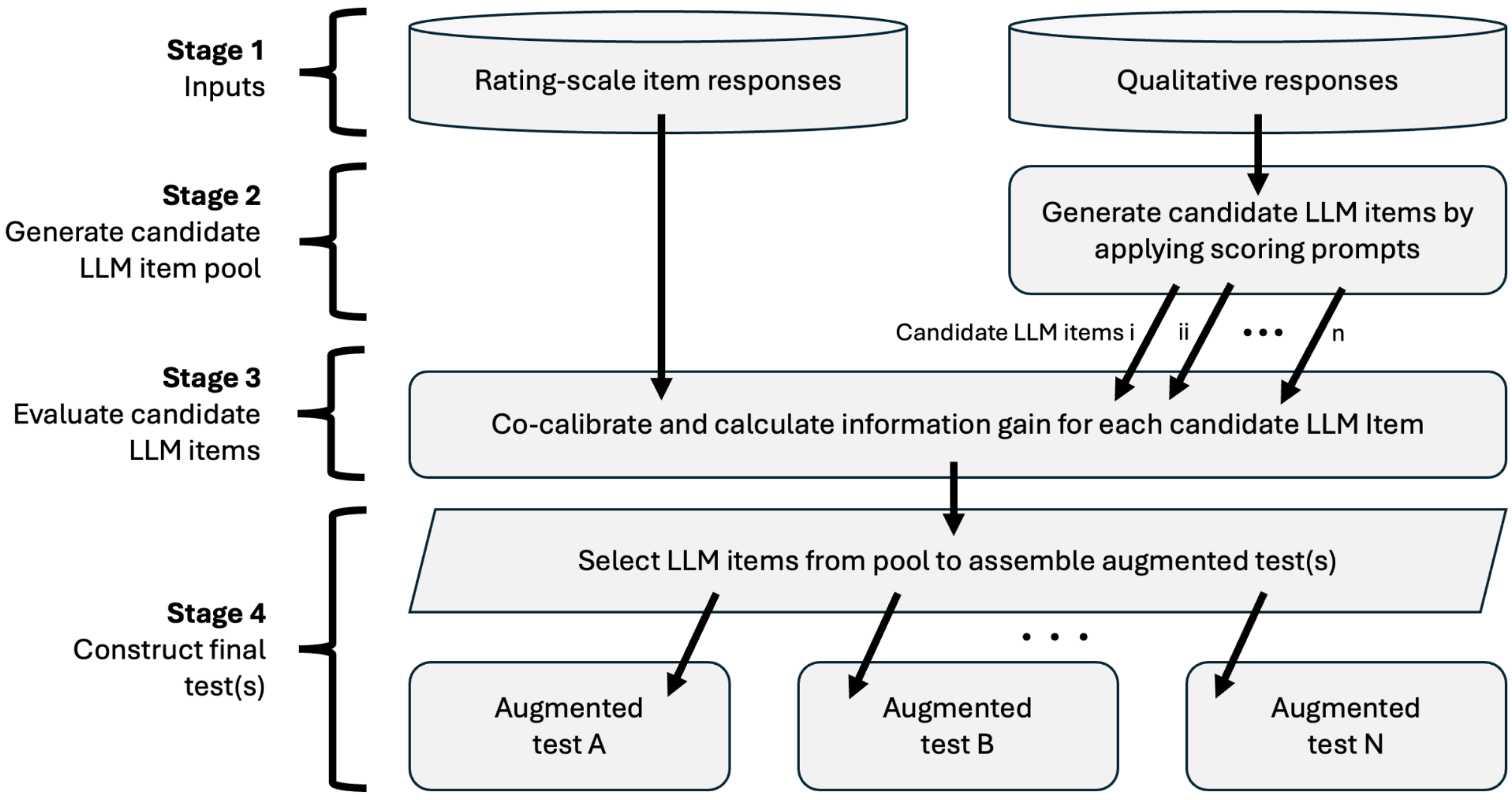

**Fig. 1.** IDS Framework for Developing Augmented Tests with LLM Items

*Note*. Schematic of the four-stage IDS framework: inputs from rating-scale responses and qualitative text (Stage 1), generation of candidate LLM items (Stage 2), information-based evaluation via co-calibration (Stage 3), and selection of items for augmented tests (Stage 4).

# Methods

This study implemented and evaluated the IDS framework, a procedure for augmenting an existing rating-scale psychological assessment with items derived from free-text responses. Using depression as a case study, we applied the framework to two data sources: (i) a real-world dataset of upper-secondary students (*Human Participant Data*) and (ii) a matched synthetic dataset generated to enable evaluation of accuracy and bias under known latent trait values (*Synthetic Data*).

For each dataset, responses were split into training and test sets. The IDS framework was applied to training data to construct two augmented depression tests ("All Texts" and "Top 5 Texts"), each comprising the baseline rating-scale items supplemented with selected LLM-derived items. These augmented tests were subsequently evaluated against the baseline test using held-out test data, through analyses of static psychometric properties and computerised adaptive testing (CAT) simulations.

## Data Sources

### Human Participant Data

Human participant data were collected from high-school students in Shaanxi Province, China, of whom 693 provided a complete set of responses comprising 13 qualitative responses (one essay and 12 sentence completion tasks), and 20 rating-scale items (19 of which were retained in the baseline test following scale purification, *Assessment Construction*). The raw rating-scale item responses were processed prior to any analysis. Items that were negatively worded were reverse-scored. Additionally, responses were converted to a binary format due to a bimodal response pattern among participants who often, and sometimes exclusively, selected the scale endpoints (0 or 10). Responses of 5 or higher were coded as 1, with all other responses coded as 0. All 693 respondents also provided a single sliding-scale value indicating their level of suicide intentionality, used later without modification to check convergent validity.

Qualitative responses were also processed before test development, with each text translated from Mandarin to English prior to scoring (using an LLM, GPT-4o-mini). To improve translation accuracy, the prompt for the model included the corresponding essay title or sentence completion stem for each piece of text. Only synthetic examples are provided for illustrative purposes (*SI 2*) – neither the original nor translated texts can be shared directly in accordance with ethical approval. Ethical approval for the collection and analysis of human participant data was granted by the Institutional Review Board of the Faculty of Psychology, Beijing Normal University (Ref: BNU202212090125). Written informed consent was obtained from all participants and their legal guardians prior to data collection.

## Synthetic Data

A synthetic dataset (*SI 2*, available in full at https://github.com/JoeMarkWatson/AES) was also generated, permitting evaluation of accuracy and bias (*Assessment Evaluation*). We first created a vector of 3,000 theta (depression) values, with a mean of 0 and standard deviation of 1. Rating-scale item responses were generated using the catR package in R (Magis & Barrada, 2017) from the 3,000 thetas in conjunction with item parameters (difficulty and discrimination), established from human participant data (*Assessment Construction*). The synthetic respondents' theta values were also used when generating responses to open-ended questions.

To generate the synthetic qualitative responses, detailed vignettes were created for each synthetic respondent (Yun et al., 2025). These included demographic information, academic background, hobbies, and a depression percentile derived from their simulated theta value. This vignette was embedded in a prompt that instructed an LLM (GPT-4o-mini) to generate corresponding essay and sentence completion responses (*SI 3*). To avoid overtly revealing an appropriate depression score through the generated text, the prompt forbade the use of any direct reference to depression. Additionally, five randomly sampled human participant responses for each qualitative task were included in the prompt as few-shot examples to promote tonal consistency between human and synthetic text.

# Procedure

## Data Partitioning

Both the human participant and synthetic datasets were partitioned into a training set (two-thirds of the data) and a test set (one-third of the data). This resulted in a training set of 2,000 and a test set of 1,000 for the synthetic data, and a training set of 462 and a test set of 231 for the human participant data. The training sets were used exclusively for assessment construction, while the test sets were held out for assessment evaluation.

## Assessment Construction

Assessment development followed the four-stage IDS framework (Fig. 1). This framework was applied independently to the real-world training set and the synthetic training set, producing two augmented tests for each data source.

### Stage 1: Collating Inputs

Our proposed framework requires responses to rating-scale items from a calibrated test and qualitative text(s) (*Data Sources*). We constructed our own calibrated rating-scale test using responses to 20 rating-scale items (*Human Participant Data*). This involved the application of Item Response Theory (IRT), a framework that models the relationship between an individual's latent trait level and their item responses. Specifically, we fit a two-parameter logistic (2PL) IRT model to the human participant training set data using the mirt package in R (Chalmers, 2012). An iterative item purification process was then conducted to promote scale quality. This resulted in the removal of one item due to local dependence with another, yielding a final set of 19 items in the calibrated baseline test.

**Stage 2: Generating the Candidate Item Pool**

In this framework, a candidate LLM item is defined as the combination of a specific qualitative task and distinct LLM prompting strategy used to score the respondent's answer. Creating a pool of candidates involved scoring the text from each of the 13 open-ended tasks (12 sentence completions and one essay), applying four distinct prompting strategies with an LLM (DeepSeek-V3.1), giving 52 (13 × 4) candidate LLM items. The four prompting strategies were designed to achieve the same goal – score the text for depression – but from varying conceptual standpoints:

- Prompt A: A normative comparison against a typical peer
- Prompt B: A rating of clinical concern
- Prompt C: Agreement with a statement about the student's emotional state
- Prompt D: Selection from a descriptive rubric of emotional content

The full text of each prompt is provided in the Supporting Information (*SI 4*).

To ensure reliability and reproducibility of LLM-generated variables, we set the model temperature to 0 for deterministic outputs (Geathers et al., 2025; Tang et al., 2024). Additionally, to permit long-term reproducibility regardless of model updates, we used a DeepSeek model that is fully downloadable. This provides a robust alternative to the practice of “model version pinning”, available for models from OpenAI (Chen et al., 2023). These controls mitigate known issues in automated scoring, such as low inter-rater reliability between models (Seßler et al., 2025), occasional hallucinations, and inconsistent adherence to scoring rubrics (Pack et al., 2024).

**Stage 3: Evaluating Candidate Items**

The goal of this stage was to quantify the psychometric information provided by each candidate LLM item, enabling later selection of scoring methods based on their information contribution. To achieve this, we iteratively co-calibrated each of the 52 candidate LLM items with the 19 original rating-scale items in a new unidimensional 2PL IRT model. Each of these models included the 19 rating-scale items plus a single candidate item. During this procedure, the parameters for the 19 rating-scale items were held constant (i.e., fixed), ensuring that the new candidate item was calibrated onto the same latent trait scale established by the baseline test, a process akin to scale-linking (Robitzsch, 2025). The value of each candidate item was then quantified by its information gain. This was calculated as the total test information of the newly augmented 20-item test minus the total test information of the 19-item baseline test, across a distribution of theta estimates (obtained by applying the baseline model to the training set). This process yielded a specific information gain value for each of the 52 candidates, allowing them to be ranked for the selection process in the next stage.

**Stage 4: Construct Augmented Tests**

In accordance with the IDS framework, we used the information values from Stage 3 to select the best-performing LLM items and construct two final augmented tests: "All Texts" and "Top 5 Texts". To build the "All Texts" test, we first retained only those candidate LLM items for which the full 1-5 response range was used by the LLM across the training set respondent sample. From this filtered set, we then selected the single most informative item for each of the 13 qualitative tasks. These 13 LLM items were combined with the 19 original rating-scale items. The resulting "All Texts" test was then calibrated by fitting a new IRT model while holding the parameters for the 19 rating-scale items fixed. The "Top 5 Texts" test was constructed as a more parsimonious version. From the 13 LLM items included in "All Texts", we retained only the five most informative for inclusion alongside the 19 rating-scale items. The 19 rating-scale and 5 LLM-derived items in the "Top 5 Texts" test were calibrated using the same parameter-fixing procedure. In this study, these two augmented tests constitute the final outputs of the implemented four-stage process; however, the framework is not restricted to this number, and researchers could generate and evaluate a larger set of augmented tests at this stage.

The construction of the augmented tests was guided by two objectives. First, we sought to avoid local dependency among items, which we addressed by adopting a "one LLM item per text" rule. This decision was informed by exploratory analyses indicating that deriving multiple LLM items from a single piece of qualitative data – an approach explored in other research (Simons et al., 2024) – could introduce problematic local dependencies. Consistent with our design choice, subsequent checks using Yen's Q3 statistic indicated no values of concern for any added LLM items. A second objective was the preservation of the original measurement construct, which was achieved by fixing the parameters of the 19 baseline rating-scale items.

Although the present study evaluates multiple augmented tests rather than selecting a single preferred instrument, the IDS framework readily supports an additional validation step for applied deployment. In such settings, researchers could partition the available data into independent training, validation, and test sets (as opposed to training and test sets only). Several candidate augmented tests could then be constructed by applying the procedures described above to the training set, and compared on the validation set using an evaluation process such as that outlined in *Assessment Evaluation*. A single test could subsequently be selected and assessed on the test set to obtain an unbiased estimate of its performance. Our objective here, however, was to examine the performance of various plausible augmented tests, and we therefore evaluated both models developed on the training set directly on the test set.

## Assessment Evaluation

The baseline test and the two augmented tests were evaluated on the held-out human participant and synthetic test sets through both an analysis of static psychometric properties and CAT simulations. We first examined the models' static properties by calculating item and test information functions to describe measurement precision across the latent trait continuum. We also calculated an Information Equivalence metric: the ratio of the information provided by an augmented test's LLM item(s) to that of a single average rating-scale item from the baseline test (calculated across $\theta$ = -2 to 2).

Simulations were conducted using CAT, a method of test administration where the next item is selected based on a respondent's current theta estimate. Our simulations were configured as follows, using custom code and the mirt package in R (Chalmers, 2012):

- Theta Estimation: After each response, theta was re-estimated using the Expected a Posteriori (EAP) method under mirt's default N(0,1) prior (Chalmers, 2012).
- Item Selection: The next item was chosen using the maximum Fisher information (MFI) criterion at the current theta estimate.
- Starting Rule: For the baseline test, the first item rating scale item was selected using the MFI criterion at the assumed population mean ($\theta = 0$). For the augmented tests, the first rating-scale item was also selected using the MFI criterion but at an initial theta estimate derived from responses to the LLM items.
- Stopping Rule: Each simulation continued until all 19 items were administered.

The performance of each model in the CAT simulations was assessed using several metrics. For precision, accuracy, and bias, the following metrics were calculated after each item administration and then averaged across all respondents in the test set:

- Theta Estimate Standard Error (SE): The average standard error of the latent trait estimates, indicating measurement precision (evaluated for both human participant and synthetic data).
- Theta Accuracy: The mean absolute error (MAE) between the estimated and true theta values (evaluated for synthetic data only).

- Theta Bias: The mean difference between the estimated and true theta values (evaluated for synthetic data only).

Convergent validity was assessed after each item administration as the sample-level association between theta estimates and an external continuous measure of suicide intentionality:

- Convergent Validity: The squared correlation ($R^2$) between theta estimates and the external measure at each test stage (evaluated for human participant data only).

Differences between baseline and augmented tests in CAT-based precision and accuracy were evaluated using repeated-measures ANOVAs, with assessment (Baseline, Top 5 Texts, All Texts) and number of administered items as within-subject factors, and respondent ID as the repeated measure. ANOVAs were conducted in R, using the afex package (Singmann et al., 2024). When omnibus effects were significant, Bonferroni-corrected pairwise contrasts were computed with Baseline as the reference condition using the emmeans R package (Lenth, 2023).

# Results

We used computerised adaptive testing (CAT) simulations to compare the performance of a baseline rating-scale-only test with two tests augmented created through the IDS framework: an “All Texts” test, which incorporated one IDS-selected LLM item per qualitative response, and a more parsimonious “Top 5 Texts” test, which included only the five most informative LLM items from the “All Texts” set (see *Procedure*). Across held-out test sets for real-world (n = 231) and synthetic (n = 1,000) data, both augmented tests showed significant improvements in precision (supporting H1). In the synthetic data, where true trait levels were known, they also showed significant gains in accuracy (supporting H2).

**Real Data Findings**

Augmented tests showed key advantages in real-world data simulations. Repeated-measures ANOVA followed by Bonferroni-corrected pairwise contrasts (see *Assessment Evaluation*) confirmed that both augmented tests produced significantly lower standard error in theta estimates (i.e., greater precision) than the baseline across all rating-scale item positions ($k = 1$–19, see Fig. 2B; and full results in *SI 5*), supporting H1. Averaged across $k = 1$–19, SE was 10.3% lower for the All Texts model and 6.9% lower for the Top 5 Texts model relative to the baseline (*SI 6*). During the early phase of the CAT ($k = 1$–10 rating-scale items), the standard error (SE) of test takers' $\theta$ estimates was reduced by an average of 12.0% for the All Texts model and 8.1% for the Top 5 Texts model relative to the baseline. Consistent with these reductions, by the end of the early phase ($k = 10$ rating-scale items), 46.3% (All Texts) and 42.4% (Top 5 Texts) of test takers achieved $SE \leq 0.3$ compared to 35.5% under the baseline test (*SI 7*). Test information increased across the $\theta$ continuum when including LLM items, particularly at higher trait levels ($\theta > 1$; Fig. 2E). The information contributed by the LLM items (Fig. 2F) was equivalent to adding up to 6.3 average rating-scale items for the All Texts model and 3.8 items for the Top 5 Texts model (calculated across $\theta = -2$ to 2). Augmented tests also showed stronger convergent validity with an external suicidality measure at every stage of the adaptive test (Fig. 2D). Mean increases in explained variance were $\Delta R^2 = .04$ (All Texts) and .03 (Top 5 Texts) during the early stages of the adaptive test (averaged across $k = 1$–10), and $\Delta R^2 = .03$ and .02 respectively across the full test ($k = 1$–19; *SI 8*).

Lastly, we examined how closely intermediate estimates approximated the baseline test's final score (Fig. 2C). When very few rating-scale items were administered, estimates from the augmented tests were approximately as close to the final 19-item baseline score as the baseline test's own early estimates. As additional rating-scale responses were collected, the baseline test's intermediate estimates converged more rapidly on its final score than did those of the augmented tests. This divergence is expected because the baseline's final score serves as its own reference standard in this comparison. Divergence is similarly observed in the synthetic data (*SI 9*), which there allows augmented test estimates to converge more closely on true θ values (Fig. 3C).

**Synthetic Data Findings**

Synthetic data simulations confirmed and extended the positive real-world data findings for tests constructed using the IDS framework. Both augmented tests again demonstrated statistically significant reductions in theta estimate standard error compared to the baseline across all rating-scale item positions (k = 1-19, see Fig. 3B; *SI 5*), further supporting H1. Averaged across k = 1–19, SE was reduced by 25.2% for the All Texts model and 20.1% for the Top 5 Texts model relative to the baseline (*SI 10*). During the early phase of the CAT (k = 1–10 rating-scale items), the standard error (SE) of respondents' θ estimates was reduced by an average of 29.6% for the All Texts model and 24.2% for the Top 5 Texts model relative to the baseline. By the end of the early phase (k = 10 rating-scale items), 60.4% (All Texts) and 57.3% (Top 5 Texts) of respondents achieved SE ≤ 0.3, compared to 34.7% under the baseline test, indicating substantially earlier high-precision measurement (*SI 11*). Test information was heightened across all theta levels for augmented tests, particularly for high-trait individuals (θ > 1; Fig. 3E). The information gain from including LLM items was equivalent to adding 16.0 average rating-scale items for the All Texts model and 10.4 items for the Top 5 Texts model (calculated across $\theta = -2$ to 2; see Fig. 3F).

Because the true $\theta$ values were known in the synthetic data, we were also able to directly evaluate estimation accuracy. Post-hoc tests confirmed significantly lower estimation error for both augmented models relative to baseline (Fig. 3C; *SI 5*), supporting H2. Mean absolute error (MAE) was reduced by 14.1% (All Texts) and 13.9% (Top 5 Texts) during the early phase ($k$ = 1–10), and by 9.7% and 10.0%, respectively, across the entire rating-scale item phase ($k$ = 1–19; *SI 12*). These improvements occurred without introducing clear systematic bias in theta estimates (Fig. 3D). Visual inspection of subgroup plots suggests that this improvement in accuracy holds across the latent trait (*SI 13*).

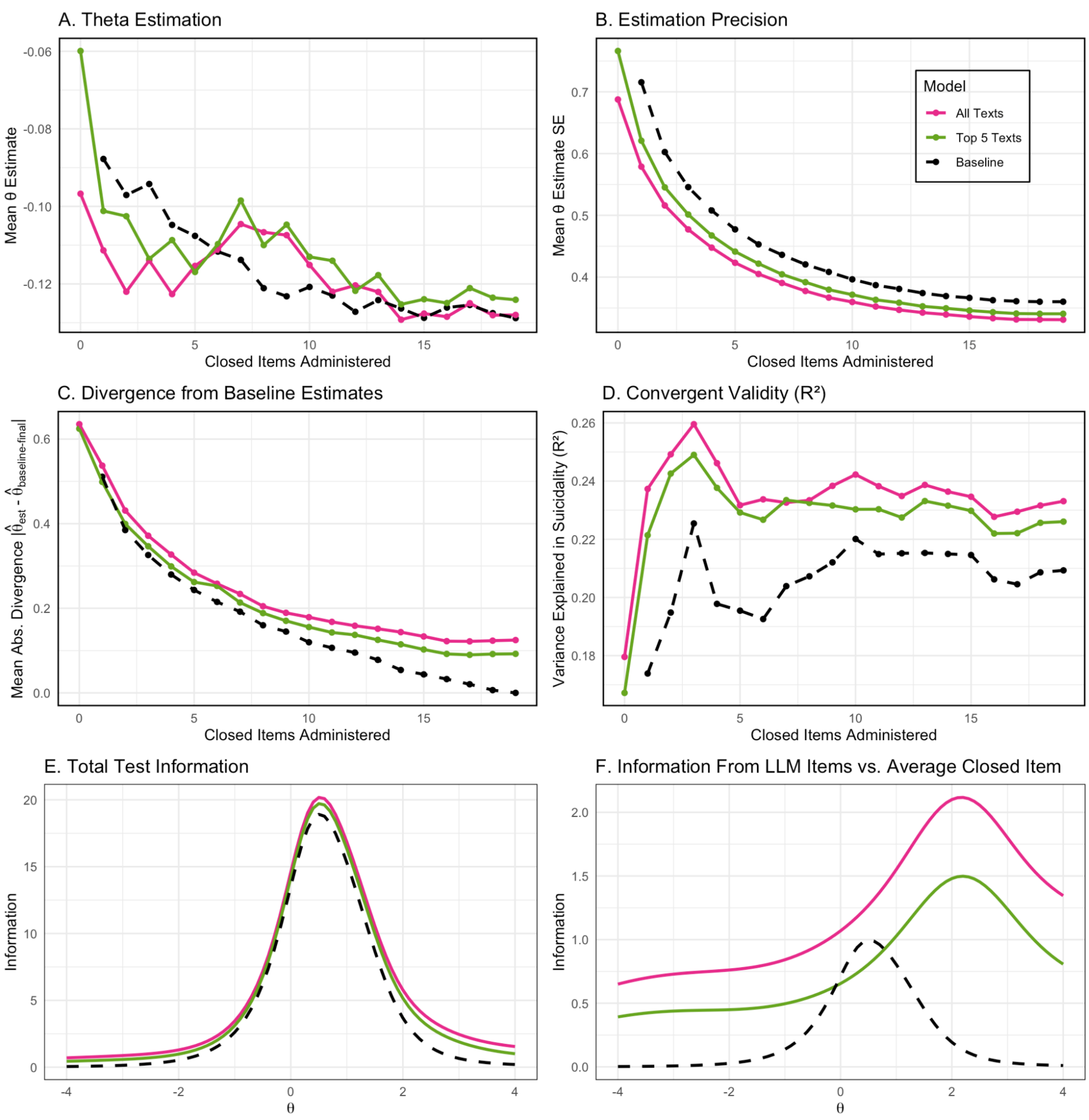

**Fig. 2.** CAT simulation Results for Real Data

*Note*. This figure displays CAT simulation results on the real-world test set (n = 231), shown as: (**A**) mean theta estimates over the course of the test; (**B**) mean standard error of theta estimates, indicating measurement precision; (**C**) mean absolute divergence of theta estimates from the final estimate of the baseline model; (**D**) convergent validity, measured as the variance explained (R-squared) in an external suicidality measure; (**E**) total test information function for each test; (**F**) information provided by the LLM items in each augmented test compared to the average information of a single rating-scale item.

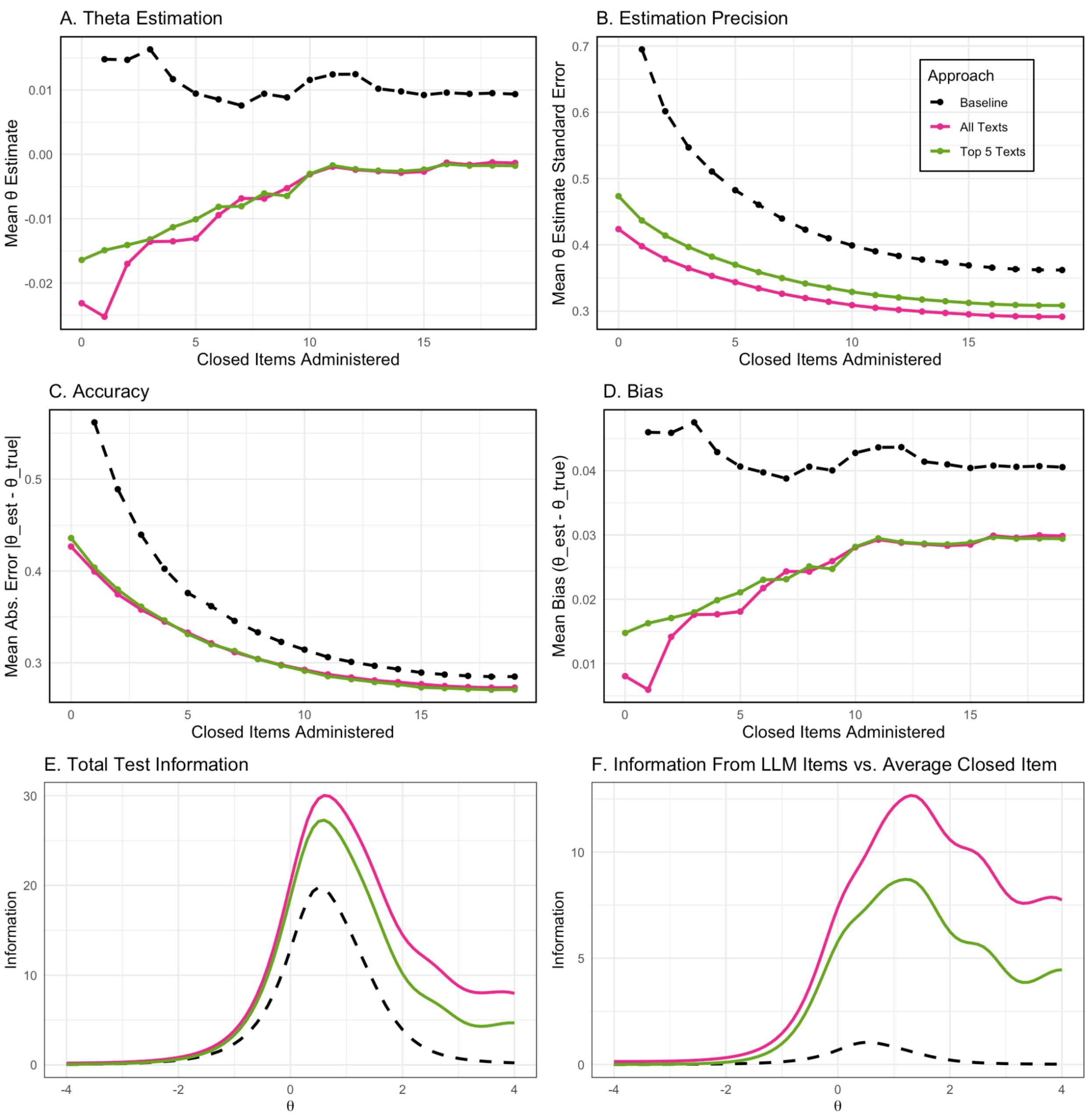

**Fig. 3.** CAT simulation Results for Synthetic Data.

*Note*. This figure presents simulation results on the synthetic test set (n = 1,000), presented as (**A**) mean theta estimates over the test; (**B**) mean standard error of theta estimates; (**C**) accuracy, measured as the mean absolute error between estimated and true theta; (**D**) bias, measured as the mean difference between estimated and true theta; (**E**) total test information function for each test; (**F**) information provided by the LLM items compared to the average information of a single rating-scale item.

# Discussion

Simulation and static comparisons between the augmented and baseline tests show that augmenting rating-scale assessments through the IDS framework can enhance measurement performance. The improved precision observed in CAT simulations (Figs. 2B, 3B) was driven by an increase in test information (Figs. 2E–F, 3E–F). While no new rating-scale responses were collected, incorporating text-derived LLM items added measurement information about the psychological trait through a process akin to extending the item bank. Further, the LLM items included in augmented tests were selected from a pool of candidates based on the test information they contributed – ensuring that only the most informative additions were retained (see *Methods*). A scale-linking procedure fixed the parameters of the original rating-scale items, meaning that added LLM items were mapped onto the same underlying depression construct. This alignment supported the observed improvements in accuracy (Fig. 3C) and low levels of bias (Fig. 3D).

CAT simulations indicate that the performance advantages of the augmented tests are most pronounced in the early stages. Because the IDS-selected LLM items are scored first, an initial approximation of a respondent's theta is generated before the rating-scale item portion of the test begins. This is evidenced by the relative accuracy of augmented tests when zero rating-scale items have been administered (Fig. 3C). This prior information, in turn, enhances the efficiency of the CAT algorithm, enabling it to select a more statistically informative initial rating-scale item for each respondent.

Instead of attempting to engineer a single, perfect prompt, the IDS framework tests multiple simple scoring strategies and empirically selects those that yield the most informative candidate LLM items. In the real-world data, different prompts (A, B, and D) emerged as the most effective means of extracting information from specific qualitative responses, whereas in the synthetic data, one prompt (B) dominated across tasks (see *SI 4* for all sets of scoring instructions). The point of these findings is not that one prompt is universally superior, but that prompt performance varies – and therefore should be evaluated empirically rather than assumed. What matters is the process: generating several candidate LLM items through easy-to-create prompts and selecting among them based on the psychometric information they provide.

The IDS framework, therefore, marks a departure from the prevailing assumption that AES should mirror human judgment. Rather than tuning models for maximum agreement with human annotations (Kermani et al., 2025; Qiu et al., 2025) or creating detailed rubrics to align the LLM's output with human preferences (Cardwell et al., 2023; Kortemeyer & Nöhl, 2025), our approach avoids such reliance on time-intensive human expertise. It substitutes this with a computational process where LLM items are chosen from multiple candidates based on the psychometric information they provide about the target trait, as opposed to their fidelity to human opinion.

The present study's limitations suggest various avenues for future research. The effectiveness of the IDS framework is contingent on the size of the candidate LLM item pool from which LLM items are selected, which was generated using only a single model (DeepSeek-V3.1) and a constrained set of LLM scoring instructions. Creating a larger candidate LLM item pool would increase the probability of discovering items that yield even greater test information. Without altering the captured qualitative text, this can be achieved by applying the same prompts to alternate LLM models. The pool could be enlarged by trialling a wider array of scoring instructions – regardless of whether these appear relevant to humans (Webson & Pavlick, 2022) – with phrasing variations likely to impact model output substantially (Chen et al., 2025; Chu et al., 2024). Additionally, more robust candidate LLM items could potentially be obtained by dedicating more computational resources at inference time to techniques such as task decomposition (Zhou et al., 2023) or ensemble methods like self-consistency, which aggregate scores from multiple model outputs (Wang et al., 2023). Furthermore, the generalisability of our findings requires examination, as our data focused on depression and were drawn from a specific demographic (Chinese high-school students). Future studies could therefore replicate this approach with diverse populations and across other psychological constructs. Finally, a particularly exciting future direction is to apply the IDS framework more broadly, leveraging the vast quantities of unstructured text now available in diverse contexts - from transcribed clinical audio to customer review data (*Introduction*).

Taken together, this study introduces the novel IDS framework for enhancing existing psychological tests by integrating information derived from qualitative responses. IDS involves a conceptual shift in automated scoring: instead of relying on pre-labelled data or complex rubrics, the process compares multiple candidate LLM items before selecting among them based on item information. Tests augmented with the chosen LLM items achieve significantly higher precision and accuracy, with these gains being most pronounced at earlier stages of CATs. These increases in precision and accuracy represent more than incremental measurement gains; our approach establishes a pathway toward more holistic assessments that can leverage the rich psychological information embedded in natural language.

# Declarations

**Funding**

This research received no specific funds or grant from any funding agency.

**Conflicts of interest**

The authors declare that they have no conflicts of interest.

**Ethics approval**

Ethical approval for the collection and analysis of human participant data was granted by the Institutional Review Board of the Faculty of Psychology, Beijing Normal University (Ref: BNU202212090125).

**Consent to participate**

Written informed consent was obtained from all participants and their legal guardians prior to data collection.

**Consent for publication**

Not applicable.

**Availability of data and materials**

The complete synthetic dataset used in this study is publicly available on GitHub: https://github.com/JoeMarkWatson/AES.

Human participant data are not publicly available due to ethical restrictions specified in the informed consent agreements with respondents.

**Code availability**

All code used for data simulation, analysis, assessment construction, and evaluation is publicly available on GitHub: https://github.com/JoeMarkWatson/AES.

**Author contributions**

Joe Watson: Conceptualisation; Methodology; Formal analysis; Writing – original draft; Writing – review and editing.

Ivan O'Connor: Formal analysis; Methodology; Writing – review and editing.

Chia-Wen Chen: Formal analysis; Methodology; Writing – review and editing.

Luning Sun: Data curation; Methodology; Writing – review and editing.

Fang Luo: Data curation.

David Stillwell: Methodology; Writing – review and editing.

# References

American National Election Studies. (2025). *ANES 2024 time series study: Pre-election and post-election survey questionnaires* [Data set]. https://electionstudies.org/wp-content/uploads/2025/08/anes_timeseries_2024_questionnaire_20240808.pdf

American Psychiatric Association. (2015). *The American Psychiatric Association practice guidelines for the psychiatric evaluation of adults*. https://doi.org/10.1176/appi.books.9780890426760.pe01

Cardwell, R., Naismith, B., LaFlair, G., & Nydick, S. (2023). *Duolingo English test: Technical manual*. Duolingo. https://doi.org/10.46999/CQNG4625

Chalmers, R. P. (2012). mirt: A multidimensional item response theory package for the R environment. *Journal of Statistical Software*, *48*(6), 1–29. https://doi.org/10.18637/jss.v048.i06

Chen, B., Zhang, Z., Langrené, N., & Zhu, S. (2025). Unleashing the potential of prompt engineering for large language models. *Patterns*, *6*(6), 101260. https://doi.org/10.1016/j.patter.2025.101260

Chen, L., Zaharia, M., & Zou, J. (2023). *How is ChatGPT's behavior changing over time?* (arXiv:2307.09009). arXiv. https://doi.org/10.48550/arXiv.2307.09009

Chu, K., Chen, Y.-P., & Nakayama, H. (2024). A better LLM evaluator for text generation: The impact of prompt output sequencing and optimization. *Proceedings of the Annual Conference of JSAI*, 1–4. https://doi.org/10.11517/pjsai.JSAI2024.0_2G5GS604

Geathers, J., Hicke, Y., Chan, C., Rajashekar, N., Sewell, J., Cornes, S., Kizilcec, R. F., & Shung, D. (2025). *Benchmarking generative AI for scoring medical student interviews in objective structured clinical examinations (OSCEs)* (arXiv:2501.13957). arXiv. https://doi.org/10.48550/arXiv.2501.13957

Ifenthaler, D. (2022). Automated essay scoring systems. In O. Zawacki-Richter & I. Jung (Eds.), *Handbook of open, distance and digital education* (pp. 1–15). Springer Nature. https://doi.org/10.1007/978-981-19-0351-9_59-1

Kermani, A., Perez-Rosas, V., & Metsis, V. (2025). *A systematic evaluation of LLM strategies for mental health text analysis: Fine-tuning vs. prompt engineering vs. RAG* (arXiv:2503.24307). arXiv. https://doi.org/10.48550/arXiv.2503.24307

Kjell, O. N. E., Kjell, K., & Schwartz, H. A. (2024). Beyond rating scales: With targeted evaluation, large language models are poised for psychological assessment. *Psychiatry Research*, *333*, 115667. https://doi.org/10.1016/j.psychres.2023.115667

Kortemeyer, G., & Nöhl, J. (2025). Assessing confidence in AI-assisted grading of physics exams through psychometrics: An exploratory study. *Physical Review Physics Education Research*, *21*(1), 010136. https://doi.org/10.1103/PhysRevPhysEducRes.21.010136

Kumar, V., Rajwat, P. S., Medda, G., Ntoutsi, E., & Recupero, D. R. (2024). Unlocking LLMs: Addressing scarce data and bias challenges in mental health and therapeutic counselling. *Proceedings of the First International Conference on Natural Language Processing and*

*Artificial Intelligence for Cyber Security*, 238–251. https://aclanthology.org/2024.nlpaics-1.26/

Lenth, R. V. (2023). *emmeans: Estimated marginal means, aka least-squares means* [Computer software]. https://CRAN.R-project.org/package=emmeans

Ludwig, S., Mayer, C., Hansen, C., Eilers, K., & Brandt, S. (2021). Automated essay scoring using transformer models. *Psych*, *3*(4), 897–915. https://doi.org/10.3390/psych3040056

Magis, D., & Barrada, J. R. (2017). Computerized adaptive testing with R: Recent updates of the package catR. *Journal of Statistical Software*, *76*(1), 1–19. https://doi.org/10.18637/jss.v076.c01

Maris, G., & Bechger, T. (2006). Scoring open ended questions. In C. R. Rao & S. Sinharay (Eds.), *Handbook of statistics* (Vol. 26, pp. 663–681). Elsevier. https://doi.org/10.1016/S0169-7161(06)26020-6

National Institute for Health and Care Excellence. (2022). *Depression in adults: Treatment and management*. https://www.nice.org.uk/guidance/ng222/resources/depression-in-adults-treatment-and-management-pdf-66143832307909

Pack, A., Barrett, A., & Escalante, J. (2024). Large language models and automated essay scoring of English language learner writing: Insights into validity and reliability. *Computers and Education: Artificial Intelligence*, *6*, 100234. https://doi.org/10.1016/j.caeai.2024.100234

Page, E. B. (1966). The imminence of... grading essays by computer. *The Phi Delta Kappan*, *47*(5), 238–243.

Peters, H., & Matz, S. C. (2024). Large language models can infer psychological dispositions of social media users. *PNAS Nexus*, *3*(6), 231. https://doi.org/10.1093/pnasnexus/pgae231

Qiu, J., Guo, D., Papini, N., Peace, N., Levinson, C. A., & Henry, T. R. (2025). *Ensemble of large language models for curated labeling and rating of free-text data* (arXiv:2501.08413). arXiv. https://doi.org/10.48550/arXiv.2501.08413

Ramesh, D., & Sanampudi, S. K. (2022). Automated essay scoring systems: A systematic literature review. *Artificial Intelligence Review*, *55*(3), 2495–2527. https://doi.org/10.1007/s10462-021-10068-2

Ravenda, F., Bahrainian, S. A., Raballo, A., Mira, A., & Kando, N. (2025). *Are LLMs effective psychological assessors? Leveraging adaptive RAG for interpretable mental health screening through psychometric practice* (arXiv:2501.00982). arXiv. https://doi.org/10.48550/arXiv.2501.00982

Robitzsch, A. (2025). Linking error estimation in fixed item parameter calibration: Theory and application in large-scale assessment studies. *Foundations*, *5*(1), 4. https://doi.org/10.3390/foundations5010004

Seßler, K., Fürstenberg, M., Bühler, B., & Kasneci, E. (2025). Can AI grade your essays? A comparative analysis of large language models and teacher ratings in multidimensional essay scoring. *Proceedings of the 15th International Learning Analytics and Knowledge Conference*, 462–472. https://doi.org/10.1145/3706468.3706527

Simons, J. J. P., Ze, W. L., Bhattacharya, P., Loh, B. S., & Gao, W. (2024). *From traces to measures: Large language models as a tool for psychological measurement from text* (arXiv:2405.07447). arXiv. https://doi.org/10.48550/arXiv.2405.07447

Singmann, H., Bolker, B., Westfall, J., Aust, F., & Ben-Shachar, M. S. (2024). *afex: Analysis of factorial experiments* [Computer software]. https://CRAN.R-project.org/package=afex

Tang, X., Chen, H., Lin, D., & Li, K. (2024). Harnessing LLMs for multi-dimensional writing assessment: Reliability and alignment with human judgments. *Heliyon*, *10*(14), e34262. https://doi.org/10.1016/j.heliyon.2024.e34262

Wang, X., Wei, J., Schuurmans, D., Le, Q., Chi, E., Narang, S., Chowdhery, A., & Zhou, D. (2023). *Self-consistency improves chain of thought reasoning in language models* (arXiv:2203.11171). arXiv. https://doi.org/10.48550/arXiv.2203.11171

Webson, A., & Pavlick, E. (2022). Do prompt-based models really understand the meaning of their prompts? In M. Carpuat, M.-C. de Marneffe, & I. V. Meza Ruiz (Eds.), *Proceedings of the 2022 Conference of the North American Chapter of the Association for Computational Linguistics: Human Language Technologies* (pp. 2300–2344). Association for Computational Linguistics. https://doi.org/10.18653/v1/2022.naacl-main.167

Yun, T., Yang, E., Safdari, M., Lee, J. H., Kumar, V. V., Mahdavi, S. S., Amar, J., Peyton, D., Aharony, R., Michaelides, A., Schneider, L., Galatzer-Levy, I., Jia, Y., Canny, J., Gretton, A., & Matarić, M. (2025). Sleepless nights, sugary days: Creating synthetic users with health conditions for realistic coaching agent interactions. *Findings of the*

*Association for Computational Linguistics: ACL 2025*, 14159–14181. https://doi.org/10.18653/v1/2025.findings-acl.729

Zhou, D., Schärli, N., Hou, L., Wei, J., Scales, N., Wang, X., Schuurmans, D., Cui, C., Bousquet, O., Le, Q., & Chi, E. (2023). *Least-to-most prompting enables complex reasoning in large language models* (arXiv:2205.10625). arXiv. https://doi.org/10.48550/arXiv.2205.10625

# Supplementary Information

## SI 1: Example American National Election Studies (ANES) Items

The main text proposes a hypothetical application of the Information-Determined Scoring (IDS) framework: creating a 'Governmental Trust' measure by integrating rating-scale and qualitative data from a legacy dataset. *SI 1* provides example items from the ANES survey that could be used for this purpose.

A unified Governmental Trust measure could be derived by applying the IDS framework to closed and open-ended items from the ANES survey. Applicable closed-ended items include:

- How often can you trust the federal government in Washington to do what is right? [Always, most of the time, about half the time, some of the time, or never / Never, some of the time, about half the time, most of the time, or always]?
- Do you think that people in government [waste a lot of the money we pay in taxes, waste some of it, or don't waste very much of it / don't waste very much of the money we pay in taxes, waste some of it, or waste a lot of it]?
- Would you say the government is pretty much run by a few big interests looking out for themselves or that it is run for the benefit of all the people? [Run by a few big interests / For the benefit of all the people]
- How many of the people running the government are corrupt? [All, most, about half, a few, or none? / None, a few, about half, most, or all?]

Applicable open-ended items include:

- 'I'd like to ask you what you think are the good and bad points about the two national parties'
- 'Is there anything in particular that you like about the Democratic party?'

# SI 2: Open-ended Tasks and Indicative Responses

*Table SI 2.1 Open-ended Task Prompts*

| Open-ended task type | Open-ended task code | Essay or sentence-completion prompt | Indicative response |
|---|---|---|---|
| Sentence completion | SC1 | Over the past week, I … | enjoyed ballet practice daily. |
| | SC2 | My parents don't know that I … | dance. |
| | SC3 | On most days, my mood … | Joyful |
| | SC4 | When I stand by the window, I … | Dream of dancing in the moonlight, feeling free. |
| | SC5 | I actually … | I actually enjoy ballet more than soccer. |
| | SC6 | My family … | Very supportive. |
| | SC7 | At night, I often … | Practice ballet. |
| | SC8 | Recently, I plan to … | Join a ballet competition and improve my skills. |
| | SC9 | I should … | Focus on my dance. |
| | SC10 | Lately, my body … | Improved. |
| | SC11 | The knife on the table can … | Slice through tough vegetables. |
| | SC12 | Next week, I plan to … | Dance. |
| Essay | E1 | Write about "My Saddest Experience". Include details such as when and where it happened, the events that unfolded, and how you felt. | The saddest experience I have had happened last summer when I had to say goodbye to my dance teacher, Mr. Thompson. He had been a huge part of my |

| | |
|---|---|
| | life for the past few years, guiding me through my ballet journey and helping me find my passion for dance. But last June, he announced that he was moving to another city for a new teaching opportunity. I felt a wave of sadness … |

*Note*. The indicative responses were all taken from a single synthetic respondent (one row of the synthetic dataset). This dataset was constructed using the method outlined in the main text (*Synthetic Data*, with supporting details in *SI 3*), where the LLM prompt included few-shot examples to promote tonal consistency with the real-world dataset.

## SI 3: Synthetic qualitative data generation

The intention of synthetic qualitative data generation was to create realistic written responses corresponding to a pre-defined vector of 3,000 theta (depression) values, with each response reflecting a unique respondent profile.

### Respondent Vignettes

Each of the 3,000 synthetic respondents was assigned a unique profile, or vignette, consisting of a randomly generated age, gender, academic major, hobby, and U.S. state of residence. These demographic data were drawn from publicly available sources and randomised for each case.

### LLM Prompt Design

The vignettes were embedded into a prompt template that instructed the LLM to generate a written response. The prompt provided the model with the respondent's profile and a qualitative description of their depression level derived from the percentile rank of their simulated theta value. We mitigated the model's tendency to directly state its depression level by including an explicit instruction to avoid keywords like 'depression' or 'depressed'.

### Methodological Considerations

The model temperature value was set to 0.7 for all generations. Full implementation details are available in the project's GitHub repository.

## SI 4: LLM Scoring Instructions

This section provides the four sets of LLM scoring instructions. The candidate item pool was generated by applying every prompt to each of the 13 qualitative text responses, consistent with Stage 2 of the IDS framework. Variables in curly braces (e.g., {writing_type}) were dynamically populated.

### Prompt A: Normative Comparison

```
You are evaluating a {writing_type} from a {country_of_origin} high
school student, written for the prompt: '{writing_prompt}'.

Compared to a typical student in their demographic, how likely is the
author to be experiencing depression or a low mood?

Use this scale:

1 = Much less likely than typical
2 = Slightly less likely than typical
3 = About as likely as typical
4 = Slightly more likely than typical
5 = Much more likely than typical

Return only the single integer score.

Student response: "{humans_response}"
```

**Prompt B: Clinical Concern**

```
You are an AI assistant performing a preliminary screening for
emotional distress. Analyze the following {writing_type} from a
{country_of_origin} high school student (prompt: '{writing_prompt}').

Rate the level of concern warranted by the student's writing regarding
potential depression or severe low mood.

Use the following scale:

1 = Very low concern
2 = Low concern
3 = Moderate concern
4 = High concern
5 = Very high concern

Respond with a single integer (1-5) and nothing else.

Student response: "{humans_response}"
```

**Prompt C: Statement Agreement**

```
You are analysing a {writing_type} by a {country_of_origin} high
school student, written for '{writing_prompt}'.
```

Indicate your level of agreement with the following statement based only on the provided text: "This student's writing suggests they are experiencing depression."

Use the scale below:

1 = Strongly Disagree

2 = Disagree

3 = Neutral / Cannot Determine

4 = Agree

5 = Strongly Agree

Return only the single integer that corresponds to your choice.

Student response: "{humans_response}"

**Prompt D: Descriptive Rubric**

Read the student's {writing_type} below. Choose the one option from the list that best describes the emotional content of the text.

1 = The text displays a clearly positive, resilient, or optimistic outlook.

2 = The text is emotionally balanced or reflects typical, everyday challenges without significant distress.

3 = The text contains noticeable elements of sadness or melancholy, typical for a student under stress.

4 = The text is dominated by a tone of emotional distress, negativity, or significant worry.

5 = The text contains explicit and pervasive themes of hopelessness, despair, or worthlessness.

Return only the number of the option you choose (1, 2, 3, 4, or 5).

The student wrote this in response to '{writing_prompt}': "{humans_response}".

## SI 5: ANOVA Results

This section reports the ANOVA results used to compare the precision and accuracy of three tests ('Baseline', 'Top 5 Texts' and 'All Texts'), through CAT simulations performed on held-out real-world and synthetic data test sets. We conducted repeated-measures ANOVAs with condition as a within-subject factor and then computed Bonferroni-corrected pairwise contrasts.

### Real-World Data

Both augmented test configurations ('Top 5 Texts' and 'All Texts') produced significantly lower theta-estimate SE than the baseline test, indicating that incorporating IDS-selected LLM items improves measurement precision (in support of H1). Moreover, the 'All Texts' condition yielded significantly higher precision than 'Top 5 Texts' (Table SI 5.1).

*Table SI 5.1 Pairwise Contrasts for Standard Error of Theta Estimates (Real-World Data)*

| Contrast | Estimate | SE | df | t ratio | p value |
|---|---|---|---|---|---|
| All Texts – Baseline | –0.04725 | 0.00302 | 230 | –15.67 | $1.19 \times 10^{-37}$ |
| Top 5 Texts – Baseline | –0.03179 | 0.00257 | 230 | –12.35 | $9.94 \times 10^{-27}$ |
| All Texts – Top 5 Texts | –0.01546 | 0.00139 | 230 | –11.09 | $1.06 \times 10^{-22}$ |

*Note.* Negative contrast estimates indicate lower SE (higher precision) in the first-named condition.

**Synthetic Data**

A repeated-measures ANOVA and Bonferroni-corrected contrasts revealed significant reductions in SE for both augmented tests, relative to the baseline. Moreover, All Texts showed substantially lower SE than Top 5 Texts, mirroring the pattern observed in the real-world data (Table SI 5.2).

*Table SI 5.2 Pairwise Contrasts for Standard Error of Theta Estimates (Synthetic Data)*

| Contrast | Estimate | SE | df | t ratio | p value |
|---|---|---|---|---|---|
| All Texts – Baseline | –0.11614 | 0.00118 | 2999 | –98.08 | $< 1 \times 10^{-300}$ |
| Top 5 Texts – Baseline | –0.09341 | 0.00108 | 2999 | –86.58 | $< 1 \times 10^{-300}$ |
| All Texts – Top 5 Texts | –0.02274 | 0.00055 | 2999 | –41.64 | $4.78 \times 10^{-299}$ |

The same procedure showed that both augmented tests (Top 5 Texts and All Texts) achieved significantly lower error than the baseline test. The small, nonsignificant difference between All Texts and Top 5 Texts suggests both augmented test configurations to yield comparable accuracy in this synthetic setting (Table SI 5.3).

*Table SI 5.3 Pairwise Contrasts for Theta Estimate Accuracy (Synthetic Data)*

| Contrast | Estimate | SE | df | t ratio | p value |
| --- | --- | --- | --- | --- | --- |
| All.Texts – Baseline | –0.03879 | 0.00374 | 2999 | –10.39 | $2.28 \times 10^{-24}$ |
| Top.5.Texts – Baseline | –0.03929 | 0.00326 | 2999 | –12.05 | $3.12 \times 10^{-32}$ |
| All.Texts – Top.5.Texts | 0.00050 | 0.00199 | 2999 | 0.25 | 1.00 |

## SI 6: SE Trajectories for Real Data

This section reports mean standard error (SE) values at each CAT stage (k = 0–19 closed items administered), together with percent reduction in SE relative to the Baseline test (n = 231). Percent reduction is calculated as:

$$100 \times \frac{SE_{Baseline} - SE_{Augmented}}{SE_{Baseline}}$$

*Table SI 6.1 Mean Standard Error (SE) and Percent Reduction in SE by CAT Stage (Real Data)*

| k | Baseline SE | All Texts SE | Top 5 Texts SE | % Reduction (All Texts) | % Reduction (Top 5 Texts) |
|---|---|---|---|---|---|
| 0 | NA | 0.69 | 0.77 | NA | NA |
| 1 | 0.72 | 0.58 | 0.62 | 19.09 | 13.23 |
| 2 | 0.6 | 0.52 | 0.55 | 14.36 | 9.49 |
| 3 | 0.55 | 0.48 | 0.5 | 12.56 | 8.13 |
| 4 | 0.51 | 0.45 | 0.47 | 11.85 | 7.96 |
| 5 | 0.48 | 0.42 | 0.44 | 11.34 | 7.54 |
| 6 | 0.45 | 0.41 | 0.42 | 10.59 | 6.89 |
| 7 | 0.44 | 0.39 | 0.4 | 10.45 | 7.23 |
| 8 | 0.42 | 0.38 | 0.39 | 10.23 | 6.85 |
| 9 | 0.41 | 0.37 | 0.38 | 10.18 | 7.04 |
| 10 | 0.4 | 0.36 | 0.37 | 9.24 | 6.27 |
| 11 | 0.39 | 0.35 | 0.36 | 8.95 | 6.14 |
| 12 | 0.38 | 0.35 | 0.36 | 8.93 | 5.87 |
| 13 | 0.37 | 0.34 | 0.35 | 8.45 | 5.73 |
| 14 | 0.37 | 0.34 | 0.35 | 8.08 | 5.34 |
| 15 | 0.37 | 0.34 | 0.35 | 8.29 | 5.59 |
| 16 | 0.36 | 0.33 | 0.34 | 8.04 | 5.35 |
| 17 | 0.36 | 0.33 | 0.34 | 8.18 | 5.52 |
| 18 | 0.36 | 0.33 | 0.34 | 8.02 | 5.41 |
| 19 | 0.36 | 0.33 | 0.34 | 8.08 | 5.44 |

*Note.* Values in the Baseline, All Texts, and Top 5 Texts columns are the mean SE across respondents at each CAT stage, after none (k = 0) to all (k = 19) closed rating-scale items have been administered. % Reduction (All Texts) and % Reduction (Top 5 Texts) report the percent reduction in mean SE relative to the Baseline at each stage k (positive values indicate lower SE than Baseline). For summary values reported in the manuscript itself (*Real Data Findings*), percent reductions are additionally averaged across stages k = 1–10 ('early phase') and k = 1–19 ('full test').

## SI 7: SE Threshold Attainment for Real Data

This section reports the proportion of respondents reaching the precision thresholds of SE ≤ .30.

*Table SI 7.1 Proportion of Respondents Reaching SE ≤ .3 by CAT Stage (Real Data)*

| k | Baseline | Top 5 Texts | All Texts |
|---:|---:|---:|---:|
| 0 | NA | 0 | 0 |
| 1 | 0 | 0 | 0 |
| 2 | 0 | 0 | 0 |
| 3 | 0 | 0 | 0 |
| 4 | 0 | 0 | 0 |
| 5 | 0 | 0 | 0 |
| 6 | 0 | 0.01 | 0.12 |
| 7 | 0.09 | 0.26 | 0.29 |
| 8 | 0.3 | 0.34 | 0.38 |
| 9 | 0.32 | 0.39 | 0.44 |
| 10 | 0.35 | 0.42 | 0.46 |
| 11 | 0.42 | 0.45 | 0.49 |
| 12 | 0.44 | 0.46 | 0.5 |
| 13 | 0.45 | 0.48 | 0.52 |
| 14 | 0.47 | 0.5 | 0.52 |
| 15 | 0.48 | 0.5 | 0.52 |
| 16 | 0.49 | 0.51 | 0.53 |
| 17 | 0.5 | 0.51 | 0.54 |
| 18 | 0.5 | 0.51 | 0.54 |
| 19 | 0.5 | 0.51 | 0.54 |

*Note.* Values correspond to the proportion of respondents with SE ≤ .3 at each stage

## SI 8: Convergent Validity Trajectories for Real Data

This section reports $R^2$ values and $\Delta R^2$ (relative to Baseline) at each CAT stage.

*Table SI 8.1 Variance Explained in Suicidality ($R^2$) and Increase Relative to Baseline ($\Delta R^2$) by CAT Stage (Real Data)*

| k | Baseline $R^2$ | All Texts $R^2$ | Top 5 Texts $R^2$ | $\Delta R^2$ (All Texts) | $\Delta R^2$ (Top 5) |
|---|---|---|---|---|---|
| 0 | NA | 0.18 | 0.17 | NA | NA |
| 1 | 0.17 | 0.24 | 0.22 | 0.06 | 0.05 |
| 2 | 0.19 | 0.25 | 0.24 | 0.05 | 0.05 |
| 3 | 0.23 | 0.26 | 0.25 | 0.03 | 0.02 |
| 4 | 0.2 | 0.25 | 0.24 | 0.05 | 0.04 |
| 5 | 0.2 | 0.23 | 0.23 | 0.04 | 0.03 |
| 6 | 0.19 | 0.23 | 0.23 | 0.04 | 0.03 |
| 7 | 0.2 | 0.23 | 0.23 | 0.03 | 0.03 |
| 8 | 0.21 | 0.23 | 0.23 | 0.03 | 0.03 |
| 9 | 0.21 | 0.24 | 0.23 | 0.03 | 0.02 |
| 10 | 0.22 | 0.24 | 0.23 | 0.02 | 0.01 |
| 11 | 0.21 | 0.24 | 0.23 | 0.02 | 0.02 |
| 12 | 0.22 | 0.23 | 0.23 | 0.02 | 0.01 |
| 13 | 0.22 | 0.24 | 0.23 | 0.02 | 0.02 |
| 14 | 0.21 | 0.24 | 0.23 | 0.02 | 0.02 |
| 15 | 0.21 | 0.23 | 0.23 | 0.02 | 0.02 |
| 16 | 0.21 | 0.23 | 0.22 | 0.02 | 0.02 |
| 17 | 0.2 | 0.23 | 0.22 | 0.02 | 0.02 |
| 18 | 0.21 | 0.23 | 0.23 | 0.02 | 0.02 |
| 19 | 0.21 | 0.23 | 0.23 | 0.02 | 0.02 |

*Note.* Values in the *Baseline*, *All Texts*, and *Top 5 Texts* columns report the variance explained in suicidality ($R^2$) at each CAT stage after none ($k = 0$) to all ($k = 19$) closed rating-scale items have been administered. The columns $\Delta R^2$ (All Texts) and $\Delta R^2$ (Top 5) indicate the increase in explained variance relative to the Baseline test at each stage k (positive values indicate stronger convergent validity).

Summary values used in the manuscript (*Real Data Findings*) are $\Delta R^2$ values additionally averaged across stages k = 1–10 ("early phase") and k = 1–19 ("full test").

# SI 9: Divergence from Baseline Estimates for Synthetic Data

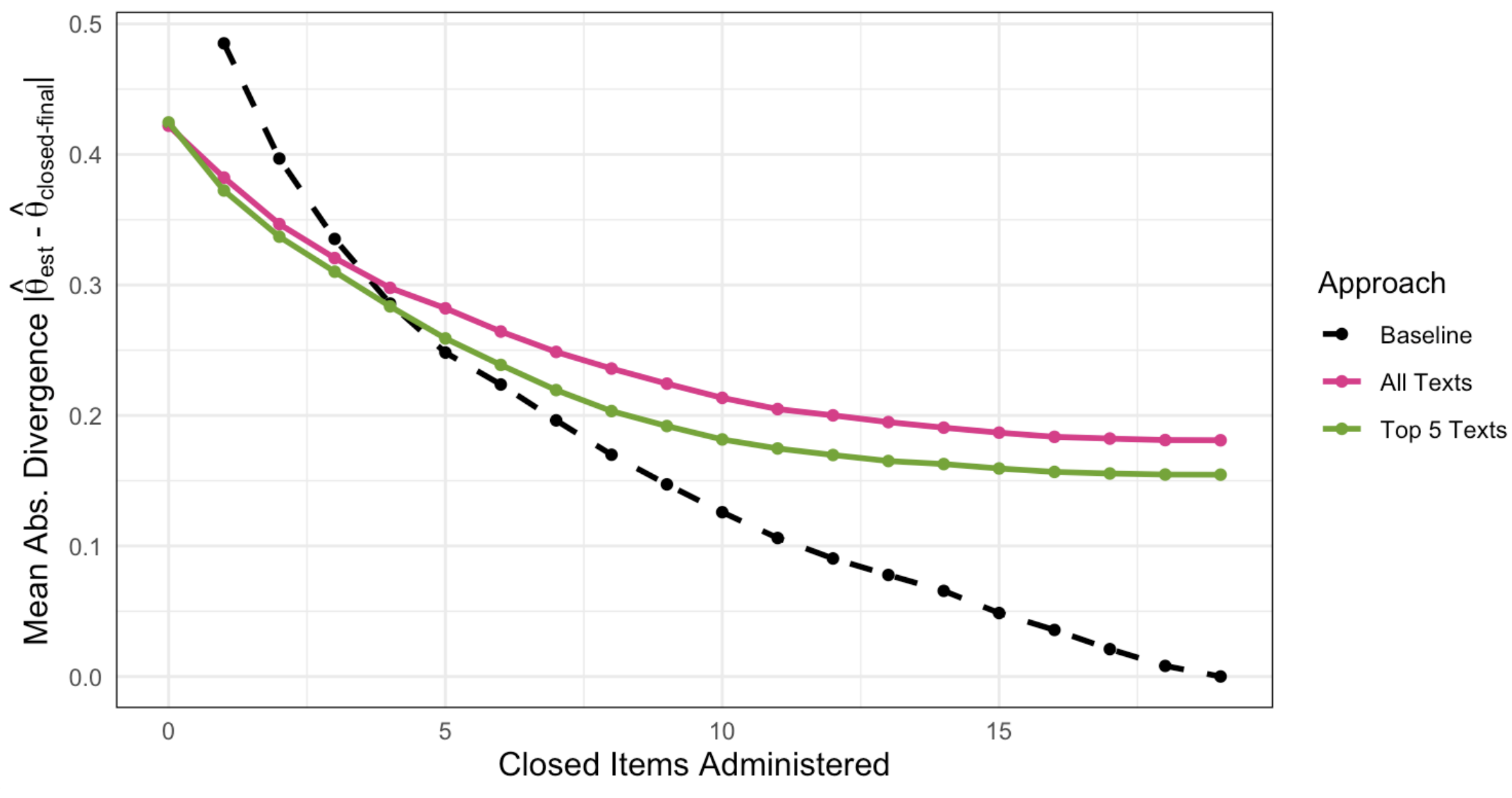

**Fig. SI 9.1.** Divergence from Baseline Estimates

*Note*. This plot provides the mean divergence between intermediate theta estimates and the final theta estimate produced by the 'Baseline' rating-scale only test.

## SI 10: SE Trajectories for Synthetic Data

This section reports mean standard error (SE) values and percent reductions relative to Baseline for synthetic simulations (n = 1,000). Percent reduction is calculated as:

$$100 \times \frac{SE_{Baseline} - SE_{Augmented}}{SE_{Baseline}}$$

*Table SI 10.1 Mean Standard Error (SE) and Percent Reduction Relative to Baseline by CAT Stage (Synthetic Data)*

| k | Baseline SE | All Texts SE | Top 5 Texts SE | % Reduction (All Texts) | % Reduction (Top 5) |
|---|---|---|---|---|---|
| 0 | NA | 0.42 | 0.47 | NA | NA |
| 1 | 0.69 | 0.4 | 0.44 | 42.74 | 37.14 |
| 2 | 0.6 | 0.38 | 0.41 | 37.06 | 31.18 |
| 3 | 0.55 | 0.36 | 0.4 | 33.33 | 27.46 |
| 4 | 0.51 | 0.35 | 0.38 | 30.85 | 25.18 |
| 5 | 0.48 | 0.34 | 0.37 | 28.77 | 23.31 |
| 6 | 0.46 | 0.33 | 0.36 | 27.4 | 22.08 |
| 7 | 0.44 | 0.33 | 0.35 | 25.81 | 20.43 |
| 8 | 0.42 | 0.32 | 0.34 | 24.4 | 19.22 |
| 9 | 0.41 | 0.31 | 0.34 | 23.33 | 18.17 |
| 10 | 0.4 | 0.31 | 0.33 | 22.57 | 17.57 |
| 11 | 0.39 | 0.31 | 0.32 | 21.85 | 16.91 |
| 12 | 0.38 | 0.3 | 0.32 | 21.22 | 16.36 |
| 13 | 0.38 | 0.3 | 0.32 | 20.77 | 15.97 |
| 14 | 0.37 | 0.3 | 0.31 | 20.38 | 15.63 |
| 15 | 0.37 | 0.3 | 0.31 | 20.02 | 15.32 |
| 16 | 0.37 | 0.29 | 0.31 | 19.8 | 15.09 |
| 17 | 0.36 | 0.29 | 0.31 | 19.57 | 14.87 |
| 18 | 0.36 | 0.29 | 0.31 | 19.49 | 14.8 |
| 19 | 0.36 | 0.29 | 0.31 | 19.46 | 14.78 |

*Note.* Mean SE values are shown for each CAT stage (k = 0–19 closed items administered) in the synthetic simulations. Percent reductions indicate the decrease in SE relative to the Baseline test at each stage (positive values indicate lower SE). For manuscript summaries (*Synthetic Data Findings*), percent reductions are additionally averaged across stages k = 1–10 ('early phase') and k = 1–19 ('full test').

## SI 11: SE Threshold Attainment for Synthetic Data

This section reports the proportion of synthetic respondents reaching SE ≤ .30 at each CAT stage (n = 1,000).

*Table SI 11.1 Proportion of Respondents Reaching SE ≤ .30 by CAT Stage (Synthetic Data)*

| k | Baseline | Top 5 Texts | All Texts |
|---|---|---|---|
| 0 | NA | 0.02 | 0.29 |
| 1 | 0 | 0.16 | 0.42 |
| 2 | 0 | 0.32 | 0.45 |
| 3 | 0 | 0.41 | 0.48 |
| 4 | 0 | 0.44 | 0.5 |
| 5 | 0 | 0.48 | 0.52 |
| 6 | 0 | 0.51 | 0.54 |
| 7 | 0.18 | 0.52 | 0.56 |
| 8 | 0.26 | 0.54 | 0.57 |
| 9 | 0.31 | 0.56 | 0.58 |
| 10 | 0.35 | 0.57 | 0.6 |
| 11 | 0.37 | 0.59 | 0.62 |
| 12 | 0.4 | 0.59 | 0.62 |
| 13 | 0.42 | 0.6 | 0.63 |
| 14 | 0.43 | 0.6 | 0.63 |
| 15 | 0.44 | 0.61 | 0.64 |
| 16 | 0.45 | 0.62 | 0.64 |
| 17 | 0.46 | 0.62 | 0.65 |
| 18 | 0.46 | 0.62 | 0.65 |
| 19 | 0.46 | 0.62 | 0.65 |

## SI 12: Accuracy Trajectories for Synthetic Data

This section reports mean absolute error (MAE) values and percent reductions relative to Baseline. Percent reduction is calculated as:

$$100 \times \frac{MAE_{Baseline} - MAE_{Augmented}}{MAE_{Baseline}}$$

*Table SI 12.1 Mean Absolute Error (MAE) and Percent Reduction Relative to Baseline by CAT Stage (Synthetic Data)*

| k | Baseline MAE | All Texts MAE | Top 5 Texts MAE | % Reduction (All Texts) | % Reduction (Top 5) |
|---|---|---|---|---|---|
| 0 | NA | 0.43 | 0.44 | NA | NA |
| 1 | 0.56 | 0.4 | 0.4 | 28.92 | 28.16 |
| 2 | 0.49 | 0.37 | 0.38 | 23.4 | 22.34 |
| 3 | 0.44 | 0.36 | 0.36 | 18.57 | 17.85 |
| 4 | 0.4 | 0.34 | 0.35 | 14.31 | 14 |
| 5 | 0.38 | 0.33 | 0.33 | 11.49 | 11.89 |
| 6 | 0.36 | 0.32 | 0.32 | 11.19 | 11.51 |
| 7 | 0.35 | 0.31 | 0.31 | 9.85 | 9.49 |
| 8 | 0.33 | 0.3 | 0.3 | 8.7 | 8.76 |
| 9 | 0.32 | 0.3 | 0.3 | 7.77 | 7.94 |
| 10 | 0.31 | 0.29 | 0.29 | 7 | 7.29 |
| 11 | 0.31 | 0.29 | 0.29 | 6.15 | 6.78 |
| 12 | 0.3 | 0.28 | 0.28 | 5.64 | 6.27 |
| 13 | 0.3 | 0.28 | 0.28 | 5.36 | 5.98 |
| 14 | 0.29 | 0.28 | 0.28 | 4.8 | 5.65 |
| 15 | 0.29 | 0.28 | 0.27 | 4.36 | 5.58 |
| 16 | 0.29 | 0.27 | 0.27 | 4.34 | 5.14 |
| 17 | 0.29 | 0.27 | 0.27 | 4.27 | 5.02 |
| 18 | 0.28 | 0.27 | 0.27 | 4.21 | 4.99 |
| 19 | 0.28 | 0.27 | 0.27 | 4.23 | 4.95 |

*Note.* Mean absolute error (MAE) values are shown for each CAT stage (k = 0–19 closed items administered) in the synthetic simulations. Percent reductions indicate the decrease in MAE relative to the Baseline test at each stage (positive values indicate improved accuracy). For manuscript summaries (Synthetic Data Findings), percent reductions are additionally averaged across stages k = 1–10 (‘early phase’) and k = 1–19 (‘full test’).

## SI 13: Accuracy and Bias by Trait Level Subgroup

Figures SI 13.1 and 13.2 break down the accuracy and bias results from the synthetic data simulations (n=1,000) for four distinct subgroups based on their true theta (θ) values.

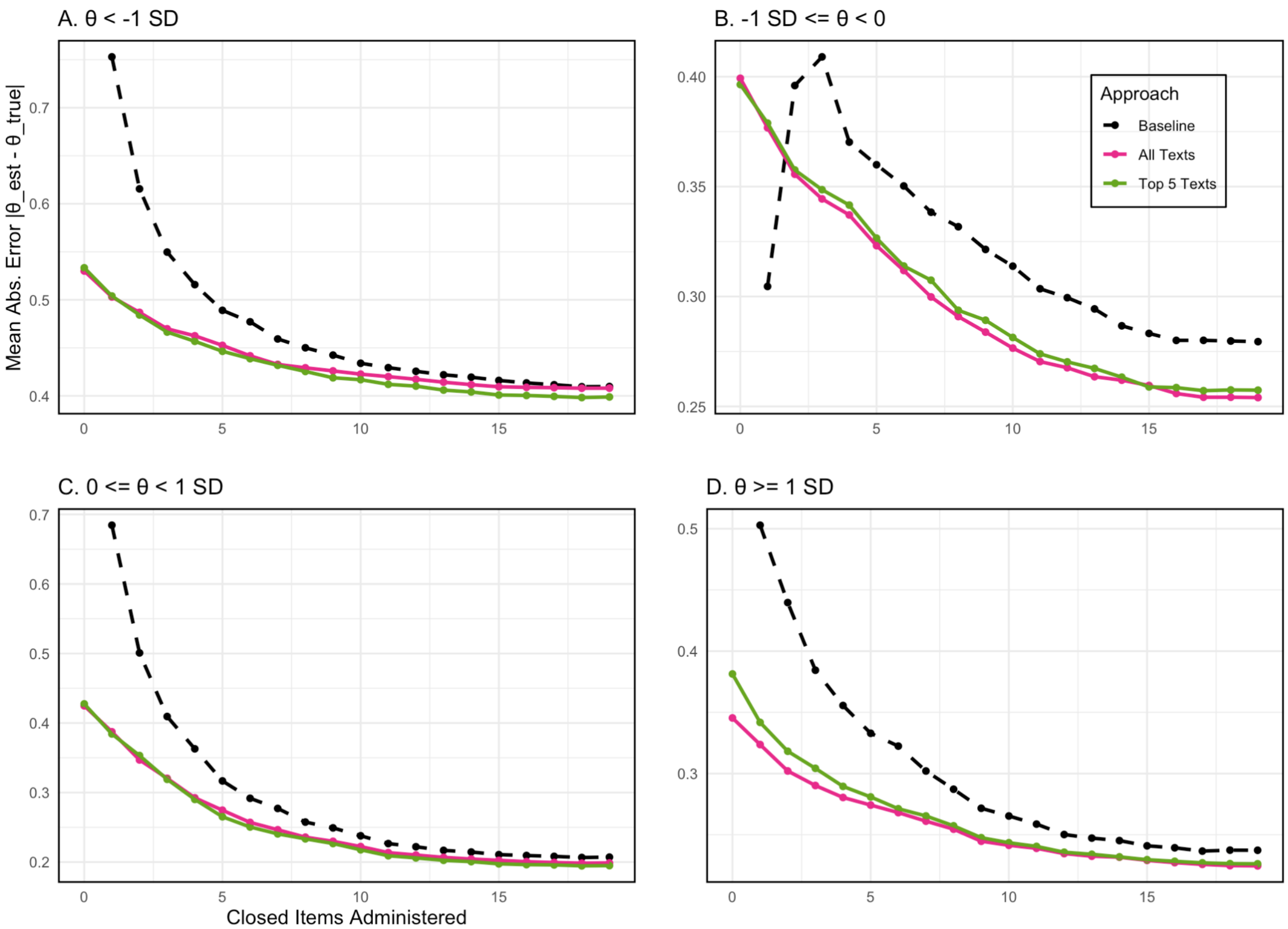

**Fig. SI 13.1.** Accuracy by Trait Level Subgroup

*Note*. The mean absolute distance between estimated and true theta is presented for respondents in four subgroups, established according to true theta level (**A-D**).

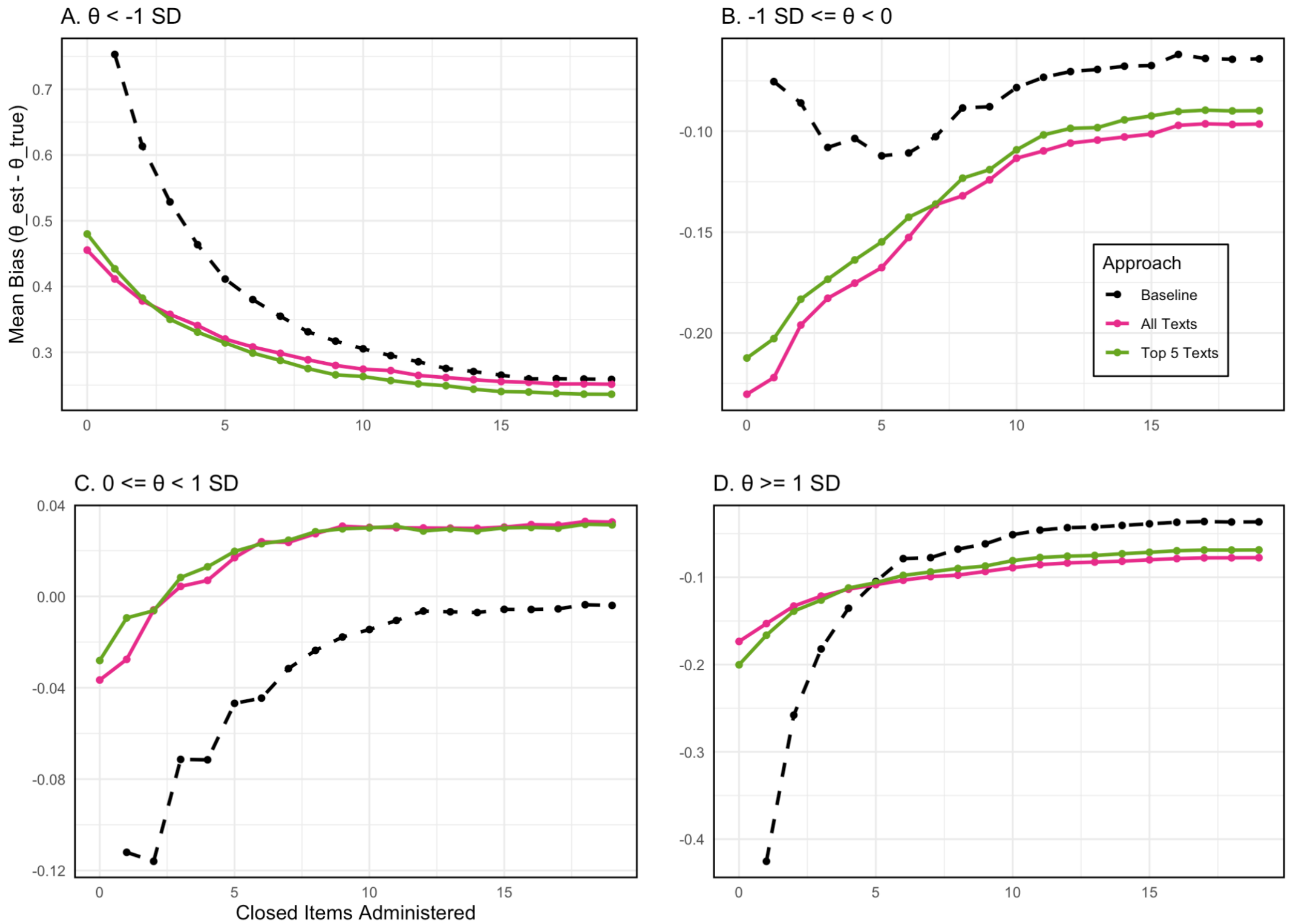

**Fig. SI 13.2.** Bias by Trait Level Subgroup

*Note*. The mean difference between estimated and true theta is given for respondents across four subgroups, established according to true theta level (**A**-**D**).